\documentclass[sigconf]{acmart}

\usepackage{tikz}
\usepackage{tabularx}
\usepackage{multirow}
\usetikzlibrary{fit, positioning}
\usetikzlibrary{arrows.meta}     
\usetikzlibrary{shapes.misc}     
\usepackage{booktabs}       
\usepackage{xcolor}       
\usepackage{graphicx}
\usepackage{subcaption} % Modern package for subfigures
\usepackage{caption}

\AtBeginDocument{%
  }

\copyrightyear{2026}
\acmYear{2026}
\setcopyright{cc}
\setcctype{by}
\acmConference[CAIN '26]{2026 IEEE/ACM 5th International Conference on AI Engineering - Software Engineering for AI}{April 12--13, 2026}{Rio de Janeiro, Brazil}
\acmBooktitle{2026 IEEE/ACM 5th International Conference on AI Engineering - Software Engineering for AI (CAIN '26), April 12--13, 2026, Rio de Janeiro, Brazil}
\acmPrice{}
\acmDOI{10.1145/3793653.3793791}
\acmISBN{979-8-4007-2475-6/2026/04}

\begin{document}

\title{The Energy Impact of Domain Model Design in Classical Planning}


\author{Ilche Georgievski}
\affiliation{%
  \institution{University of Stuttgart}
  \city{Stuttgart}
  \country{Germany}}
\email{ilche.georgievski@iaas.uni-stuttgart.de}

\author{Serhat Tekin}
\affiliation{%
  \institution{University of Stuttgart}
  \city{Stuttgart}
  \country{Germany}}
\email{serhattekin@outlook.de}

\author{Marco Aiello}
\affiliation{%
  \institution{University of Stuttgart}
  \city{Stuttgart}
  \country{Germany}}
\email{marco.aiello@iaas.uni-stuttgart.de}

\renewcommand{\shortauthors}{Georgievski et al.}

\begin{abstract}
AI research has traditionally prioritised algorithmic performance, such as optimising accuracy in machine learning or runtime in automated planning. The emerging paradigm of Green AI challenges this by recognising energy consumption as a critical performance dimension. Despite the high computational demands of automated planning, its energy efficiency has received little attention. This gap is particularly salient given the modular planning structure, in which domain models are specified independently of algorithms. On the other hand, this separation also enables systematic analysis of energy usage through domain model design. We empirically investigate how domain model characteristics affect the energy consumption of classical planners. We introduce a domain model configuration framework that enables controlled variation of features, such as element ordering, action arity, and dead-end states. Using five benchmark domains and five state-of-the-art planners, we analyse energy and runtime impacts across 32 domain variants per benchmark. Results demonstrate that domain-level modifications produce measurable energy differences across planners, with energy consumption not always correlating with runtime.
\end{abstract}




\maketitle

\section{Introduction}

AI systems offer unprecedented capabilities but also raise growing environmental concerns due to their substantial computational and energy demands. This issue is particularly relevant to the optimisation of algorithmic performance, where improvements are traditionally pursued with respect to accuracy, runtime, or solution quality. In response, Green AI research has gained momentum~\cite{schwartz2020:green-ai,verdecchia2023:green-ai}, studying the energy and carbon costs of AI systems, primarily in the context of machine learning~\cite{georgiou2022:green-ai,yarally2023:green-ai}. In contrast, AI planning remains unexplored from an energy perspective, despite its intensive computation and broad applicability~\cite{georgievski2025:green-ai-planning}. 

Classical planners are symbolic reasoning engines that automatically generate sequences of actions transforming a given initial state into a state that achieves specified goals. A sequence of actions forms a \textit{plan}, where multiple valid plans may exist that together constitute the \textit{solution space}. AI planners operate through computational pipelines involving grounding, preprocessing, heuristic computation, and search. A defining characteristic is their \textit{domain-independent architecture} in which planning algorithms are decoupled from application-specific knowledge. Application dynamics are instead encoded in a \textit{domain model}, typically specified in the Planning Domain Definition Language (PDDL)~\cite{mcdermott1998:pddl}. Domain models formally describe actions, predicates, and state transitions~\cite{mccluskey2004:knowledge-formulation}. Domain models are reused across multiple problem instances, enabling flexibility and reuse but also shift significant planner's computational behaviour to domain model design decisions (see~\cite{kambhampati2007:modelling}).

Despite their formal semantics, domain models permit considerable freedom in specification. Domain modellers must make numerous design decisions, such as action granularity, predicate structure and ordering, ordering of preconditions, and encoding style. Such design decisions are often guided by experience-driven heuristics rather than systematic principles~\cite{kambhampati2007:modelling,mccluskey2004:knowledge-formulation}. As a result, multiple semantically equivalent but syntactically distinct \emph{domain model configurations} may coexist. Extensive prior work has shown that such choices can dramatically influence the performance of planners: variations in action ordering and representation can lead to significant performance differences~\cite{howe2002:benchmark-assessment,riddle2011:domain-representation,vallati2015:dmc,vallati2021:dmc}, while seemingly benign changes, such as introducing dummy objects, can severely degrade performance~\cite{vallati2019:domain-engineering}. These effects are comparable in magnitude to differences caused by inherent domain structure~\cite{hoffmann2005:domain-structure,long2003:ipc}.

The impact of domain model design is further amplified by the iterative nature of modelling. Domain models evolve through repeated cycles of modification, validation, and debugging, each requiring full planner execution~\cite{silva2020:pre-post-design,georgievski2023:dlaip,georgievski2023:sdlc-aip}. Once deployed, the same domain models continue to shape computational behaviour during operational use, as every planner invocation reuses the computational pipeline that makes extensive use of the domain models. Consequently, domain model design choices influence not only runtime and coverage but also the cumulative computational footprint across the software development life cycle.

To date, however, research on domain modelling and planner performance has focused on traditional metrics such as runtime, plan quality, and coverage. In parallel, Green AI studies show that design decisions can cause energy consumption to vary by orders of magnitude in machine learning~\cite{yarally2023:green-ai,alizadeh2024:green-ai,yuan2024:nlp-energy,duran2024:green-ml}. For AI planning, the impact of domain model design on energy consumption remains unexplored, leaving a critical gap for energy-aware AI planning.

We address this gap by exploring how domain model design decisions impact the energy consumption of classical planners. Our contributions include: (i) a systematic framework for characterising domain model configurations; (ii) experiments across classical planners and benchmark domains from the International Planning Competition (IPC)~\cite{ipc}; and (iii) the first empirical evidence linking domain model configuration to energy consumption in AI planning. 

\section{Domain Model Configuration Framework}\label{sec:framework}

The \textit{Domain Model Configuration Framework} provides a structured approach to examining how design choices in domain models affect both the energy consumption and performance of AI planning systems. In simple terms, the framework helps researchers and practitioners systematically vary and test aspects of a domain model to understand their impact on planner behaviour. Building on prior work~\cite{howe2002:benchmark-assessment,vallati2015:dmc,vallati2019:domain-engineering,vallati2021:dmc}, the framework organises the process of identifying, classifying, and transforming domain features to produce controlled model variants for empirical evaluation. Conceptually, it serves as an intermediary between an original domain and its systematically derived variants. The framework comprises four components: \textit{Feature Identification}, \textit{Feature Taxonomy}, \textit{Transformation Principles}, and \textit{Configuration Mechanisms}.

\paragraph{Feature Identification.}
This component defines the inclusion criteria for features that can be meaningfully modified within a domain model. We postulate that candidate features must (i) be explicitly represented in the syntax or semantics of the planning language (e.g., PDDL); (ii) allow modification across domains; (iii) be non-redundant with other selected features; and (iv) remain expressible within the syntactic scope of the chosen language level (e.g., PDDL~1.2). These criteria ensure that only operationally manipulable and semantically valid features are considered.

\paragraph{Feature Taxonomy.}
Identified features are organised into three categories representing principal dimensions of domain model variation: (1) \textbf{Syntactic Structure Choices (SSC)}, referring to the organisation and ordering of domain elements, such as predicate declarations, action definitions, preconditions, and effects; (2) \textbf{Modelling Redundancy Choices (MRC)}, referring to the inclusion of semantically inert constructs, such as redundant predicates, parameters, or actions; and (3) \textbf{Task Design Choices (TDC)}, referring to modifications affecting task solvability or difficulty, such as introducing dead-end states or removing goal-achieving effects. These categories define the configuration space in which representational design decisions can be systematically studied.

\paragraph{Transformation Principles.}
Normative rules govern valid transformations, ensuring that generated variants are interpretable, comparable, and experimentally meaningful. Each transformation must (i) conform to the formal syntax and semantics of the planning language (\textit{formal validity}); (ii) preserve the original solution or exhibit explicitly defined deviations (\textit{controlled deviation}); and (iii) be applicable consistently across domains (\textit{systematic uniformity}).

\paragraph{Configuration Mechanisms.}
The framework’s operational layer defines how transformations are implemented within each taxonomy category, see Table~\ref{tab:mechanisms}. SSC mechanisms reorder domain elements by criteria (e.g., usage frequency or alphabetical order), preserving semantics while varying syntax. MRC mechanisms introduce representational redundancy by adding inert elements, inflating arity, duplicating inapplicable actions, or inserting neutral predicates, thereby maintaining solvability but increasing representational overhead. TDC mechanisms alter task solvability by introducing dead-end states or blocking goal achievement through non-removable or alternating predicates.

\begin{table*}[t]
\centering
\caption{Configuration mechanisms for domain model variant generation.}
\label{tab:mechanisms}
\begin{tabularx}{\textwidth}{p{2.4cm} l X p{3cm}}
\toprule
\textbf{Category} & \textbf{Mechanism} & \textbf{Transformation Principle} & \textbf{Impact} \\ 
\midrule
\multirow{10}{2.4cm}{\textbf{Syntactic Structure Choices (SSC)}} 
 & SSC--PDU$_{1,2}$ & Reorder predicate declarations by usage frequency in preconditions/effects (descending (D)/ascending (A)). & \multirow{10}{3cm}{No change in semantics, search space, or solution space. Space exploration may change, potentially affecting computation.} \\
 & SSC--PDA$_{1,2}$ & Reorder predicate declarations alphabetically (A-Z/Z-A). & \\
 & SSC--OEF$_{1,2}$ & Sort actions by number of effect literals (D/A). & \\
 & SSC--ONE$_{1,2}$ & Sort actions by number of negative effects (D/A). & \\
 & SSC--OPR$_{1,2}$ & Sort actions by number of preconditions (D/A). & \\
 & SSC--OPA$_{1,2}$ & Sort actions by number of parameters (D/A). & \\
 & SSC--ORA$_{1,2}$ & Sort actions by ratio of effects to preconditions (D/A). & \\
 & SSC--OAN$_{1,2}$ & Sort actions alphabetically by action name. & \\
 & SSC--PRA$_{1,2}$ & Sort preconditions within actions alphabetically. & \\
 & SSC--EFA$_{1,2}$ & Sort effect literals alphabetically within actions. & \\ 
\midrule
\multirow{7}{2.4cm}{\textbf{Modelling Redundancy Choices (MRC)}} 
 & MRC--ROB & Add 10\% additional dummy objects to the problem file (unused in any predicate or goal). & \multirow{7}{3cm}{No change in semantics or solution space. Search space and computation may change.} \\
 & MRC--RPD & Insert 10\% dummy predicates declared without parameters. & \\
 & MRC--RPA & Increase predicate arity by adding a dummy parameter. &  \\
 & MRC--ROP & Duplicate an action; make it inapplicable via contradictory preconditions. & \\
 & MRC--ROA & Add an unused dummy parameter to all action definitions. & \\
 & MRC--RPR & Add one dummy predicate to every action precondition (disjunctively). & \\
 & MRC--REF & Add a dummy predicate to action effects (positive and negative literals). & \\
\midrule
\multirow{4}{2.4cm}{\textbf{Task Design Choices (TDC)}} 
 & TDC--DEF & Duplicate an action and remove one goal-achieving effect. & \multirow{4}{3cm}{Changes semantics, solution space, and search-space structure.} \\ 
 & TDC--RPD & Introduce an unremovable dummy predicate that marks dead-end states. &  \\ 
 & TDC--APD & Duplicate an action; remove a goal-relevant effect; alternate predicates. &  \\ 
 & TDC--COP & Compose two actions such that goal-relevant effects are removed. &  \\ 
\bottomrule
\end{tabularx}
\end{table*}

\section{Experimental Evaluation}

We present a preliminary empirical study on \textit{how domain model configuration choices influence the energy consumption of classical AI planners across different planners and domains}. We first outline the methodology and experimental setup, followed by the results. A full account of the methodology, experimental setup, and tooling is available in the reproducibility artefact.\footnote{https://github.com/PlanX-Universe/energy-classical-planning-dmc}

\subsection{Methodology and Setup}

Our empirical study follows the PLANERGYM framework for energy measurement in AI planning systems~\cite{georgievski2025:green-ai-planning} and evaluates five state-of-the-art planners across five benchmark domains. The selected planners represent two dominant planning frameworks. Three planners are based on Fast Downward (FD): Fast Downward Stone Soup Agile (FDSSA)~\cite{buechner2023:fdssa}, Cerberus Agile (CA)~\cite{katz2018:cerberus}, and DALAI Agile (DALAIA)~\cite{buechner2023:dalaia}; and two based on LAPKT: Approximate Novelty Search Tarski (ANST)~\cite{singh2023:anst} and Forward–Backward Novelty Search (FBNS)~\cite{singh2023:fbns}. The benchmark domains are Barman, Blocks World, Gripper, Thoughtful, and Ricochet Robots, selected from various IPCs. For each domain, 32 domain model variants (1 original + 20 SSC, 7 MRC, and 4 TDC variants) were generated using our framework, and 10 problem instances were considered.

All experiments were conducted on a dedicated machine equipped with an Intel Core i7-1065G7 processor (4 cores, 8 threads) and 32~GB of RAM, running Ubuntu 24.04 LTS (kernel 6.8.0-60-generic). To minimise background interference and ensure stable energy measurements, the system was booted into a minimal environment without desktop services or network daemons. Energy consumption was measured at the CPU package level using Intel RAPL counters accessed via \texttt{pyRAPL}, within isolated containerised executions.

Each planner run was restricted to a single CPU core, 8~GB of memory, and a 5-minute wall-clock timeout, consistent with the IPC Agile Track settings. Each planner-domain-instance combination was executed 30 times to ensure statistical robustness. Because some configurations time out, we consider only runs that terminated within 5 minutes with a planning outcome (plan or failure).

\subsection{Results}

\textbf{Syntactic structure choices have a limited impact on the energy consumption}. Across most planner-domain combinations, energy use patterns remain stable, with deviations primarily confined to LAPKT-based planners in selected domains and most pronounced for variants that modify action or effect orderings.

Table~\ref{tab:ssc-summary} summarises the results across all domains and planners. \textbf{FD-based planners show remarkable stability}, with mean energy deviations below 1\% and near-perfect Pearson correlations across domain variants. FDSSA is the most consistent across all five domains, followed by CA and DALAIA. Isolated outliers, such as SSC-OPR1 in Gripper with FDSSA, appear to stem from instance-level difficulty rather than systematic effects of syntactic ordering.

\begin{table}[t]
\centering
\caption{SSC results. $\mu$ for mean energy, $\sigma$ for standard deviation, $t$ for mean duration, and $r$ for Pearson correlation. In bold are notable deviations from the original domain.}
\label{tab:ssc-summary}
\small
\begin{tabular}{@{}p{1.2cm}lcccc@{}}
\toprule
\textbf{Domain} & \textbf{Planner} & \textbf{$\mu$ [J]} & \textbf{$\sigma$ [J]} & \textbf{$t$ [s]} & \textbf{$r$} \\ 
\midrule
\multirow{5}{*}{\textbf{Gripper}} 
 & FDSSA  & 12.5--12.6 & 1.41--1.50 & 0.87 & 0.98--1.00 \\ 
 & CA     & 5.65--5.72 & 0.34--0.38 & 0.39 & 0.88--1.00 \\ 
 & DALAIA & 11.8--11.9 & 1.28--1.35 & 0.83 & 0.99--1.00 \\ 
 & FBNS   & 4.73--4.77 & 0.46--0.51 & 0.34 & 0.99--1.00 \\ 
 & ANST   & 7.96--8.03 & 0.42--0.51 & 0.58 & 0.94--1.00 \\ [3pt] 
\textbf{Blocks World} 
 & FDSSA  & 42.7--43.0 & 33.9--34.4 & 2.91--2.93 & 1.00 \\ [3pt] 
\multirow{3}{*}{\textbf{Barman}} 
 & FDSSA  & 45.0--45.6 & 11.5--11.7 & 3.06--3.11 & 1.00 \\ 
 & FBNS   & \textbf{51.7--114} & 15.5--186 & \textbf{3.80--9.00} & 0.04--1.00 \\ 
 & ANST   & \textbf{25.7--153} & 3.77--264 & 2.10--12.50 & \textbf{-0.19--1.00} \\ [3pt] 
\multirow{2}{*}{\textbf{Thoughtful}} 
 & FDSSA  & 257--261 & 451--459 & 18.4--18.5 & 1.00 \\ 
 & ANST   & \textbf{7.09--30.3} & 0.02--52.9 & \textbf{0.53--2.55} & \textbf{-0.68--1.00} \\ [3pt] 
\multirow{5}{1cm}{\textbf{Ricochet Robots}} 
 & FDSSA  & 455--466 & 1220--1253 & 34.5--35.6 & 1.00 \\ 
 & CA     & 358--361 & 754--759 & 27.6--27.7 & 1.00 \\ 
 & DALAIA & 128--130 & 314--321 & 10.2--10.4 & 1.00 \\ 
 & FBNS   & \textbf{41.3--51.9} & 34.1--41.8 & 3.00--3.85 & 0.79--1.00 \\ 
 & ANST   & 13.7--13.9 & 4.57--4.98 & 1.01--1.03 & 0.99--1.00 \\ 
\bottomrule
\end{tabular}
\end{table}

\textbf{In contrast, LAPKT-based planners show domain-dependent sensitivity}. ANST remains stable in Gripper and Ricochet Robots but diverges markedly in Barman and Thoughtful. In Barman, effect and action reordering lead to \textbf{large energy deviations} (25.7-153~J), with correlations dropping to near-zero or negative values. For example, SSC-OEF1 requires 153~J, whereas SSC-OEF2 requires only 39.6~J, a 3.8-fold difference. Predicate and precondition permutations, however, remain tightly correlated with the original domain. In Thoughtful, all action ordering variants fail within approximately 0.5 seconds, while predicate and precondition variants remain stable (30.2-30.3~J, perfect correlation). Effect permutations yield slight improvements, with SSC-EFA2 reducing energy to 27.22~J.

FBNS follows a similar but weaker pattern. Energy usage remains stable in Gripper, but in Barman, \textbf{effect sorting variants substantially reduce energy use} (SSC-EFA1: 63.8~J; SSC-EFA2: 51.7~J) compared to other variants (113-114~J), accompanied by runtime reductions from roughly 9~s to 3.8-4.8~s. SSC-EFA2 also deviates in Ricochet Robots, showing consistent sensitivity to effect ordering.

Considering the effects of redundant modelling constructs, we observe that energy use patterns reveal selective sensitivity across planners and domains. While most redundancy mechanisms produce modest effects, \textbf{MRC-ROA consistently increases energy consumption} across nearly all planner-domain combinations, with the severity of this effect varying substantially by domain structure. 

Table~\ref{tab:mrc-summary} summarises the results. For most configurations, mean energy consumption, runtime, and Pearson correlations remain close to the original domain. Both FD- and LAPKT-based planners are largely resilient to redundancy, indicating that minor representational inefficiencies have limited energy impact. In contrast, \textbf{MRC-ROA dominates as the primary source of inefficiency}, increasing mean energy consumption by roughly 2$\times$ to over 12$\times$ and showing high variability (up to 1400~J standard deviation).

\begin{table}[t]
\centering
\caption{MRC results. Metrics as in Table~\ref{tab:ssc-summary}.}
\label{tab:mrc-summary}
\small
\begin{tabular}{@{}p{1.2cm}lcccc@{}}
\toprule
\textbf{Domain} & \textbf{Planner} & \textbf{$\mu$ [J]} & \textbf{$\sigma$ [J]} & \textbf{$t$ [s]} & \textbf{$r$}  \\ 
\midrule
\multirow{5}{*}{\textbf{Gripper}}
 & FDSSA  & \textbf{12.5--22.2} & 1.42--5.84 & 0.95--1.51 & 0.98--1.00 \\ 
 & CA     & \textbf{5.64--12.3} & 0.35--6.62 & 0.39--0.78 & 0.93--1.00 \\ 
 & DALAIA & \textbf{11.9--21.0} & 1.30--5.61 & 0.84--1.43 & 0.99--1.00 \\ 
 & FBNS   & \textbf{4.73--13.1} & 0.46--9.64 & 0.35--0.92 & 0.97--1.00 \\ 
 & ANST   & \textbf{7.98--12.6} & 0.43--6.22 & 0.58--0.94 & 0.94--1.00 \\ 
\textbf{Blocks World}
 & FDSSA  & \textbf{42.8--487} & 34.0--1133 & 3.00--39.0 & 0.89--1.00 \\ 
\multirow{3}{*}{\textbf{Barman}}
 & FDSSA  & \textbf{45.1--90.2} & 11.2--24.6 & 3.10--6.40 & 0.82--1.00 \\ 
 & FBNS   & \textbf{48.2--585} & 13.8--311 & 4.00--44.0 & 0.29--1.00 \\ 
 & ANST   & \textbf{30.5--478} & 6.21--773 & 2.50--42.5 & 0.04--1.00 \\ 
\multirow{2}{*}{\textbf{Thoughtful}}
 & FDSSA  & \textbf{256--1487} & 445--1417 & 19.0--118 & 0.89--1.00 \\ 
 & ANST   & \textbf{27.2--380} & 39.8--526 & 2.50--34.5 & 0.92--1.00 \\ 
\multirow{5}{1cm}{\textbf{Ricochet Robots}}
 & FDSSA  & 456--473 & 1229--1256 & 33.6--35.0 & 1.00 \\ 
 & CA     & \textbf{347--467} & 721--947 & 27.6--36.5 & 1.00 \\ 
 & DALAIA & \textbf{129--163} & 316--402 & 10.2--12.7 & 1.00 \\ 
 & FBNS   & \textbf{39.6--178} & 29.6--187 & 3.00--14.0 & 0.79--1.00 \\ 
 & ANST   & \textbf{13.8--53.2} & 4.64--66.4 & 1.00--4.20 & 0.73--1.00 \\ 
\bottomrule
\end{tabular}
\end{table}

MRC-ROA affects FDSSA across all five domains, with severity increasing with domain complexity. In Gripper, mean energy rises from 12.5~J to 22.2~J (+77.6\%) with a slight correlation drop. In Blocks World, energy increases more than 11-fold (487~J), with substantial runtime and variance growth, and one instance that solves in the original domain fails under MRC-ROA. Energy use again doubles in Barman and increases more than 5-fold in Thoughtful, where several instances fail. Ricochet Robots is the exception: energy remains tightly bounded (456-473~J), with MRC-ROA slightly below baseline and MRC-RPA highest at about 15~J above baseline.

The other FD-based planners show smaller deviations. In Gripper, CA shows increased energy for MRC-ROA, MRC-RPA, and MRC-RPR, while other variants remain near baseline. In Ricochet Robots, MRC-ROA again stands out, though the deviation is minor. DALAIA follows a similar pattern, with MRC-ROA nearly doubling energy use in Gripper and causing modest increases in Ricochet Robots.

The same pattern appears for the LAPKT-based planners. \textbf{ANST shows the strongest MRC-ROA sensitivity}, particularly in Barman and Thoughtful, where mean energy increases 12-fold or more, with longer runtimes and reduced correlations; two outliers further amplify this effect without causing timeouts. In Barman, MRC-REF (35.7~J, $r=0.04$) and MRC-ROP (30.5~J, $r=0.15$) exhibit \textbf{correlation degradation despite near-baseline energy}, suggesting altered search trajectories linked to effect and action changes. In Thoughtful, MRC-REF slightly reduces energy (27.2~J), while MRC-ROP increases it moderately (43.0~J). In Gripper, MRC-ROA is again the only notable deviation, and in Ricochet Robots it triples energy use (13.8~J to 53.2~J), with correlation dropping to 0.73 and runtime increasing by over three seconds. Other variants (MRC-RPA, MRC-REF, MRC-ROP) show modest increases (15.0-16.9~J).

FBNS follows this pattern: in Barman, \textbf{MRC-ROA yields the highest mean energy use} (a 5-fold increase), while MRC-RPA and MRC-REF reduce energy use and runtime to less than half of the original. A comparable effect appears in Ricochet Robots, where MRC-ROA increases energy use 4-fold and MRC-RPA again yields lower consumption. In Gripper, MRC-ROA increases energy use from 4.7~J to 13.1~J, while MRC-ROP shows a smaller effect.

In most cases, TDC variants produce energy consumption and runtime close to baseline levels, but some TDC variants cause large deviations. As shown in Table~\ref{tab:tdc-summary}, different dead-end mechanisms affect different planner–domain combinations: some domains remain resilient, while others exhibit strong sensitivity.

\begin{table}[t]
\centering
\caption{TDC results. Metrics as in Table~\ref{tab:ssc-summary}.}
\label{tab:tdc-summary}
\small
\begin{tabular}{@{}p{1.2cm}lcccc@{}}
\toprule
\textbf{Domain} & \textbf{Planner} & \textbf{$\mu$ [J]} & \textbf{$\sigma$ [J]} & \textbf{$t$ [s]} & \textbf{$r$}  \\ 
\midrule
\multirow{2}{*}{\textbf{Gripper}}
 & FDSSA  & 13.2--14.3 & 1.74--2.27 & 0.91--0.98 & 0.97--1.00 \\ 
 & CA     & 5.93--15.2 & 0.44--12.4 & 0.40--1.09 & 0.92--0.98 \\
 & DALAIA & \textbf{12.45--13.28} & 1.53--1.89 & 0.87--0.91 & 0.99--1.00 \\
 & FBNS   & 4.89--5.06 & 0.56--0.74 & 0.35--0.36 & 0.98--0.99 \\
 & ANST   & 8.08--8.27 & 0.48--0.73 & 0.59--0.60 & 0.96--0.99 \\ [3pt]
\multirow{1}{1cm}{\textbf{Blocks World}}
 & FDSSA  & \textbf{58.5--1465} & 50.9--1898 & 5.00--105 & 0.22--0.79 \\ [3pt]
\multirow{3}{*}{\textbf{Barman}}
 & FDSSA  & 44.3--47.9 & 10.7--14.9 & 3.01--3.28 & 0.93--0.97 \\ 
 & FBNS   & 52.1--119 & 16.3--185 & 3.95--9.30 & -0.14--0.48 \\ 
 & ANST   & 27.9--103.5 & 7.86--210 & 2.20--8.90 & -0.09--0.40 \\ [3pt] 
\multirow{2}{*}{\textbf{Thoughtful}}
 & FDSSA  & \textbf{167--273} & 382--484 & 11.7--19.2 & 0.77--1.00 \\ 
 & ANST   & 31.2--173 & 38.1--194 & 2.30--16.0 & 0.45--0.97 \\ [3pt] 
\multirow{5}{1cm}{\textbf{Ricochet Robots}}
 & FDSSA  & \textbf{211--490} & 296--1262 & 16.0--35.1 & 0.81--1.00 \\ 
 & CA     & \textbf{358--1013} & 748--1997 & 28.0--63.5 & 0.90--1.00 \\ 
 & DALAIA & \textbf{134--348} & 332--979 & 10.5--28.0 & 1.00 \\ 
 & FBNS   & 51.5--101 & 40.7--113 & 3.95--7.50 & 0.82--0.99 \\ 
 & ANST   & 6.14--28.6 & 0.03--27.1 & 0.50--2.25 & -0.38--0.94 \\ 
\bottomrule
\end{tabular}
\end{table}

FDSSA shows strong domain-dependent sensitivity to task design modifications. In Blocks World, \textbf{TDC-DEF produces the largest deviation observed}, increasing energy more than 30-fold, extending runtime beyond 100~s, and reducing correlation to $r=0.69$. In Thoughtful, the effect reverses: TDC-DEF reduces energy from 260~J to 168~J while maintaining a high correlation ($r=0.78$). In Barman and Gripper, effects remain minor, with small deviations and near-constant correlation. Ricochet Robots shows mixed behaviour: TDC-COP reduces energy from 458~J to 211~J with lower runtime ($\sim$17~s) and $r=0.81$, while TDC-RPD increases energy moderately to 490~J. Other variants remain close to baseline.

Other FD-based planners produced results only for Gripper and Ricochet Robots. In Gripper, CA shows modest sensitivity: TDC-APD, TDC-COP, and TDC-DEF increase mean energy use by roughly two to three times relative to the original domain, with correlations of $r=0.92$. In Ricochet Robots, both planners show substantial effects via different mechanisms. CA is most affected by TDC-APD, with energy use rising from 350~J to 1013~J while correlation remains near-perfect. DALAIA shows a similar pattern under TDC-COP, increasing energy use from 127~J to 348~J, maintaining perfect correlation, and extending runtime from 10~s to over 27.5~s.

LAPKT-based planners show a similar pattern. For ANST, all TDC variants remain stable in Gripper, with no meaningful energy or correlation changes, while clear increases appear in other domains. In Barman, TDC-COP nearly triples energy consumption (37.3~J to 104~J), with correlation dropping to $r=0.40$ and runtime increasing from about 3~s to 9~s. Thoughtful shows an even stronger effect, with TDC-DEF raising energy use from 30.3~J to 173~J. Ricochet Robots shows a weaker but similar pattern, with TDC-RPD increasing energy use from 13.8~J to 28.6~J. FBNS shows moderate variation in Barman across TDC variants (52–119~J), with correlations ranging from -0.09 to 0.48, indicating altered search trajectories without a single dominant mechanism. In Ricochet Robots, TDC-COP produces the largest deviation, increasing energy from 42.6~J to 101~J while maintaining near-perfect correlation.

\section{Discussion}

The results reveal that energy use in classical planners is generally stable under minor representational changes but sensitive to specific modelling and task-level mechanisms that likely affect grounding effort or search-space structure. Although the magnitude of these effects varies by planner and domain, a pattern emerges: energy deviations occur when configurations change the number or distribution of reachable states, increase the grounding load, or introduce conditions that affect heuristic evaluation. Energy efficiency thus reflects structural interactions among domain model configurations, planners, and domain topologies, rather than runtime alone.

Three main patterns are observed. First, \textbf{SSC variations are largely benign}. Reordering predicates, preconditions, or actions produces only minor energy and runtime changes, particularly for FD-based planners, consistent with effective preprocessing and normalisation. Exceptions arise mainly in LAPKT-based planners, where action or effect permutations sometimes disrupt heuristic evaluation, likely due to order-dependent data structures or search behaviour. Second, \textbf{MRC mechanisms have the strongest impact}. Modelling redundancy, especially redundant action arity, produces systematic and often large energy increases across planners, frequently by an order of magnitude. These effects are consistent with increased grounding effort, search-space size, and memory use, and typically preserve moderate-to-high correlation with baseline behaviour, indicating uniform computational scaling rather than difficulty reordering. Third, \textbf{TDC variations show domain- and planner-dependent effects}. While most variants remain close to baseline, some introduce extreme deviations when dead ends interact with domain constraints, amplifying grounding complexity or pruning search space portions. FD-based planners appear particularly exposed, while LAPKT-based planners show greater resilience. TDC mechanisms thus occupy an intermediate position, sometimes increasing energy sharply and sometimes reducing it.% when constraints simplify search.

\textbf{Planner architecture explains much of the observed variability}. FD-based planners show pronounced sensitivity to redundancy and dead ends, consistent with architectures that front-load computation through grounding and preprocessing, while remaining robust to syntactic variation. LAPKT-based planners, in contrast, are more resilient to redundancy but occasionally sensitive to structural ordering, particularly where novelty evaluation depends on action or effect order. These differences point to \textbf{architecture-specific vulnerabilities} rather than uniform planner behaviour.

Domain topology further shapes energy behavior. \textbf{Blocks World and Barman emerge as highly sensitive domains}, where rich relational structures magnify modelling and task-level changes into disproportionate search expansions. Thoughtful shows moderate sensitivity regarding solvability thresholds, while \textbf{Gripper and Ricochet Robots show exceptional stability}, suggesting compact domains maintain predictable energy use when configured.

\textbf{These findings carry practical implications}. For domain modellers, representational efficiency is critical: avoiding redundant action parameters and carefully validating task design choices can substantially reduce energy consumption without changing planners. For planner developers, the results highlight opportunities for energy-aware preprocessing, redundancy detection, and the use of order-independent data structures. More broadly, \textbf{energy emerges as a performance metric that exposes inefficiencies invisible through runtime alone}.

Finally, the evaluation focuses on classical planning, IPC-style benchmarks, and a subset of planners, with partial coverage due to timeouts. Although experiments were conducted under controlled conditions with repeated executions, results may not generalise to other planning paradigms, architectures, or deployment environments. Furthermore, our interpretations here are based on observed empirical patterns and established characteristics of planner architectures; direct measurements and attribution to specific internal components (e.g., grounding or heuristic evaluation) is beyond the scope of this study.

\section{Conclusion}\label{sec:conclusion}

We examined how domain model configuration affects the energy consumption of classical planners through a systematic empirical study covering three configuration categories, five planners, and five benchmark domains. The results show that syntactic orderings are largely energy-neutral except in LAPKT-based planners, where they can induce up to fourfold variation; redundant action arity consistently increases energy consumption across planners by factors of 2-12 and can lead to failures; and dead-end introduction produces effects ranging from negligible to catastrophic, driven primarily by domain structure rather than planner architecture. These findings demonstrate that energy efficiency in classical planning is a structural property arising from the interaction among domain model configuration, planner design, and domain topology, and cannot be inferred solely from runtime. The Domain Model Configuration Framework provides a systematic means to characterise these sensitivities and supports future research. 

Future work should develop predictive models that estimate energy costs from static domain model properties, enabling energy-aware modelling without exhaustive empirical testing. Further investigation is needed to explain planner-specific sensitivities to action and effect ordering, the resilience of certain domains (e.g., Ricochet Robots) to action-arity penalties, and the differing behaviour of novelty-based versus goal-directed heuristics. Extending the analysis to richer PDDL features, additional planning frameworks, and real-world domains would strengthen generalisability, while studying combined design choices may reveal new results.

\bibliographystyle{ACM-Reference-Format}
\bibliography{lit}


\end{document}